\documentclass[9pt, conference]{IEEEtran}
\IEEEoverridecommandlockouts

\usepackage{cite}
\usepackage{amsmath,amssymb,amsfonts}
\usepackage{algorithmic}
\usepackage{graphicx}
\usepackage{booktabs}
\usepackage{textcomp}
\usepackage{xcolor}
\usepackage[acronym]{glossaries}
\usepackage{multirow}
\usepackage{hyperref}
\usepackage{cleveref}
\usepackage{subfig}
\usepackage{inconsolata}
\usepackage{relsize}
\def\BibTeX{{\rm B\kern-.05em{\sc i\kern-.025em b}\kern-.08em
    T\kern-.1667em\lower.7ex\hbox{E}\kern-.125emX}}

\makeglossaries
\newacronym{prm}{PrM}{Prediction Module}
\newacronym{pom}{PoM}{Policy Module}
\newacronym{iism}{IISM}{Inter-Iteration Selection Module}
\newacronym{mt}{MT}{Mockturtle}
\newacronym{sa}{SA}{Step-Ahead}
\newacronym{mig}{MIG}{Major-Inverter-Gate}
\newacronym{dse}{DSE}{Design Space Exploration}
\newacronym{rmse}{RMSE}{Root Mean Square Error}
\newacronym{sota}{SoTA}{State-of-The-Art}
\newacronym{aig}{AIG}{And-Inverter Graph}
\newacronym{qor}{QoR}{Quality of Results}

\begin{document}

\title{The Art of Beating the Odds with\\Predictor-Guided Random Design Space Exploration
\thanks{
    *Felix Arnold and Maxence Bouvier are co-first authors.
    }
}
\IEEEaftertitletext{\vspace{-2\baselineskip}}

\author{\IEEEauthorblockN{Felix Arnold*}
\IEEEauthorblockA{Huawei, Switzerland}
\and
\IEEEauthorblockN{Maxence Bouvier*}
\IEEEauthorblockA{Huawei, Switzerland}
\and
\IEEEauthorblockN{Ryan Amaudruz}
\IEEEauthorblockA{Huawei, Switzerland}
\and
\IEEEauthorblockN{Renzo Andri}
\IEEEauthorblockA{Huawei, Switzerland}
\and
\IEEEauthorblockN{Lukas Cavigelli}
\IEEEauthorblockA{Huawei, Switzerland}
}

\maketitle

\begin{abstract}
This work introduces an innovative method for improving combinational digital circuits through random exploration in MIG-based synthesis.
High-quality circuits are crucial for performance, power, and cost, making this a critical area of active research.
Our approach incorporates next-state prediction and iterative selection, significantly accelerating the synthesis process.
This novel method achieves up to 14\texttimes\ synthesis speedup and up to 20.94\% better MIG minimization on the EPFL Combinational Benchmark Suite compared to state-of-the-art techniques.
We further explore various predictor models and show that increased prediction accuracy does not guarantee an equivalent increase in synthesis quality of results or speedup, observing that randomness remains a desirable factor.
\end{abstract}

\vspace{-10pt}

\section{Introduction}
In recent years, substantial efforts have been made to transition the synthesis of combinational circuits from proprietary to open-source EDA tools. Still, these tools lag behind commercial alternatives in terms of area, power, timing---particularly for well-known arithmetic blocks.
All-in-one tools, such as YOSYS \cite{yosys}, often perform minimal combinational module minimization, while more specialized software like ABC \cite{abc_tool} and \acrfull{mt} \cite{mt_tool} require high expertise to configure synthesis scripts and algorithms.
S.-Y. Lee et al. \cite{dac24_mt} recently introduced a random-search-based \acrfull{dse} framework for \acrfull{mig} minimization, achieving remarkable improvements over the \acrfull{sota}.
However, random search typically involves numerous trial-and-error iterations to find an effective algorithmic sequence.
F. Faez et al. \cite{predictive} have explored prediction-based synthesis and demonstrated \acrfull{qor} improvements and speedups, but they did not report the absolute minimal circuit size achieved.
We propose a method built upon the framework of \cite{dac24_mt} that accelerates the \acrshort{dse} by incorporating next-state prediction and iterative selection.

\section{Method}
\begin{figure*}
    \centering
    \includegraphics[width=0.95\textwidth]{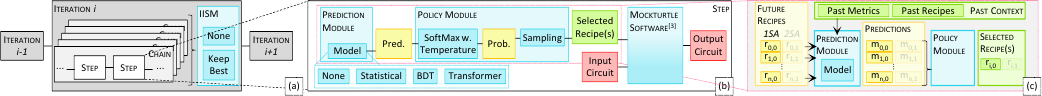}
    \caption{
    Overview of the proposed prediction-based \acrshort{dse} framework.
    (a) Multiple parallel chains and \acrshort{iism}.
    (b) Unitary steps with recipe selection and \acrshort{mt} recipe processing.
    (c) Detailed view of recipe performance prediction and recipe selection by the \acrshort{prm} and \acrshort{pom}.
    }
    \label{fig:flowy_overview}
    \vspace{-10pt}
\end{figure*}

The original system \cite{dac24_mt} decomposes the full optimization process into a sequence of actions (steps) performed by the \acrfull{mt} software \cite{mt_tool}.
At each step, one of thirty possible recipes, drawn from six distinct synthesis scripts with varying parameters, is randomly selected and applied to the circuit.
Our method introduces three modules that collaborate to optimize the synthesis \acrshort{qor}.
The \acrfull{prm} and \acrfull{pom} optimize the sequence of recipes (see Fig. \ref{fig:flowy_overview}b) to achieve faster circuit minimization.
The \acrfull{iism} periodically restarts the synthesis from the best circuit found so far (Fig. \ref{fig:flowy_overview}a).

\textbf{Recipe Selection Process:}
A \textit{recipe} is an atomic action that the \acrshort{mt} tool performs during a unitary step.
The \acrshort{prm} predicts the resulting circuit size after applying all possible recipes to the current circuit.
Based on the predictions, the \acrshort{pom} selects the recipe that \acrshort{mt} should apply next.
In this work, we evaluate several \acrshort{prm} models, including statistical and decoder-based causal auto-regressive Transformer \cite{transformer}.
In the \acrshort{pom}, the predicted metrics associated with each recipe are passed through a SoftMax function with temperature to obtain probabilities.
These probabilities are used to sample the next action, as illustrated in Fig. \ref{fig:flowy_overview}c.

\textbf{Data Collection:}
To train the predictor models, we collect data by applying uniformly sampled recipes for a certain number of steps.
At each step, we record the recipe index along with the resulting size-related metrics (LUT6, transistor, and MIG node counts, etc.).

\textbf{Inter-Iteration Selection Process:}
After a fixed number of steps, the \acrshort{iism} selects the best circuit.
All chains are restarted from the selected design, initiating a new iteration.
Parallel chains start from the same circuit and proceed with an individual sequence of recipes.

\section{Experiments}
In a first step, we choose to optimize the ALU control unit (\texttt{control.v}) from the EPFL Benchmark \cite{epfl_benchmark}, as its small size allowed for fast iteration.
In this section, the target metric used is the number of transistors, as reported by YOSYS \cite{yosys} after a technology independent \texttt{abc} step.
Data collection is done with 500 runs, each with a single chain of 5000 steps.
Unless otherwise stated, for all data points of each experiment presented below, we perform 200 runs, each with 1000 steps across iterations and chains.
The real compute time can differ, as the \acrshort{prm} and recipe execution durations can vary.
A single \acrshort{mt} script takes 0.065\,s on average (0.27\,s including metrics extraction) on a 32-core 2.8\,GHz x86\_64 server CPU. 

\begin{figure}
    \centering
    \includegraphics[width=1.00\linewidth]{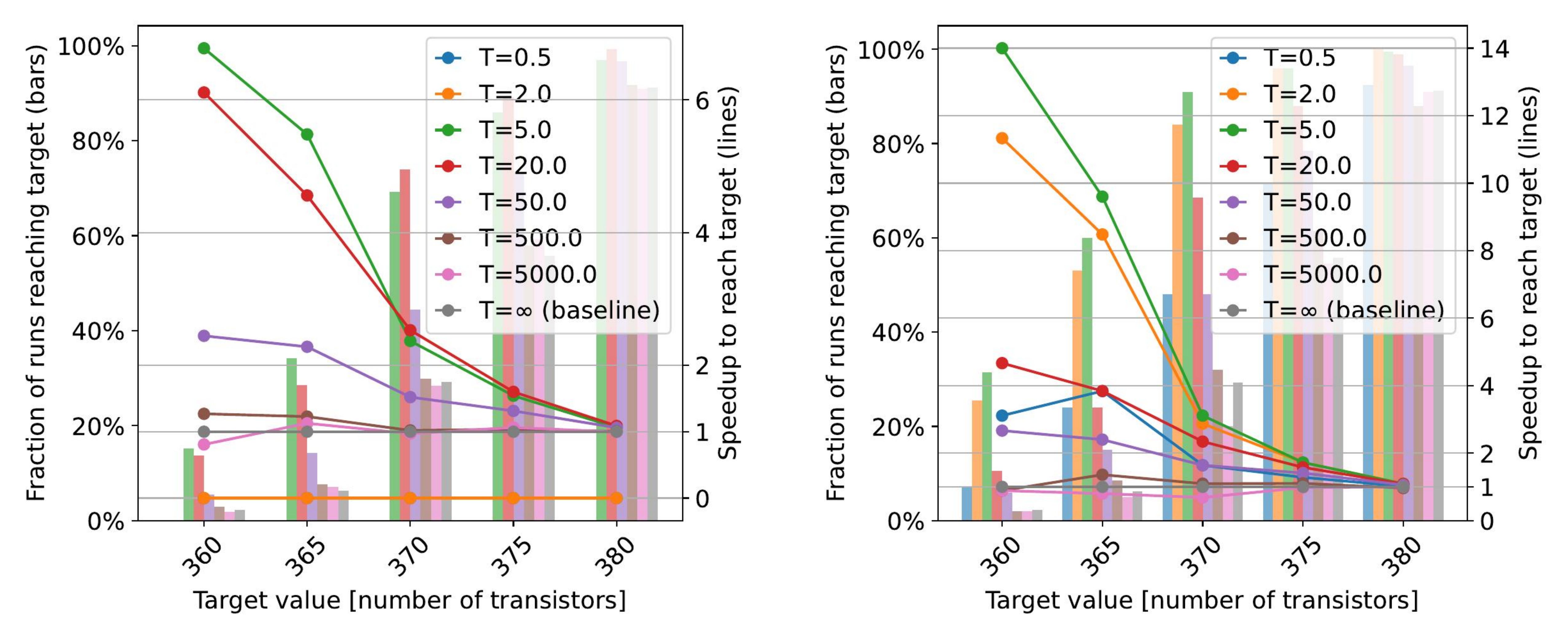}
    \caption{Effect of the temperature on the speedup factor for various target values for \texttt{ctrl.v}. Statistical model. a) 1SA, b) 2SA.}
    \label{fig:speedup_combined}
    \vspace{-5pt}
\end{figure}

\textbf{Prediction Accuracy:}
We investigate several prediction models:
1) The statistical model predicts transistor variations based on the mean change observed for each recipe in the training set; 
2) Other models incorporate metrics derived from previous circuits, with context lengths ranging up to 20 past states for the transformer-based model.
Table \ref{tab:prm_prediction} summarizes the \acrfull{rmse} achieved on a validation set by different models across various configurations (lower is better).
We note that higher prediction accuracy does not necessarily lead to higher \acrshort{qor}.

\begin{table}[hbp]
\centering
\caption{RMSE of the \acrshort{prm}, speedup and minimum size achieved.}
\label{tab:prm_prediction}
\begin{tabular}{lrrrr}
\toprule
    \textbf{Configuration}&\textbf{\acrshort{rmse}}& \textbf{Speedup@365} & \textbf{Min. Trans.}\\
    \midrule
    Uniform (Baseline)&-&1.00&354\\
    \midrule
    Statistical, 1SA&19.4&5.48&\textbf{344}\\
    MLP, 1SA &\textbf{17.5}&\textbf{5.71}&348\\
    \midrule
    Statistical, 2SA&23.6&\textbf{9.60}&346\\
    Transformer, 2SA&\textbf{18.0}&9.00&\textbf{340}\\
    \midrule
    Statistical, 1SA $+$ \acrshort{iism}&19.2&10.60&340\\
\bottomrule
\end{tabular}
\end{table}

\textbf{Temperature Effect:}
Higher SoftMax temperature results in a more uniform distribution during sampling, whereas lower temperatures favor the predicted top-performing recipes.
To investigate the impact of the parameter on the \acrshort{qor} we perform a sweep (see Fig. \ref{fig:speedup_combined}a).
We see that there is an optimum around $T=5$, where significantly better results can be achieved compared to the uniform distribution (baseline).
This suggests selecting top-performing recipes improves \acrshort{qor}, but keeping some degree of randomness is essential.

\textbf{Speedup:}
We define synthesis speed as the average number of runs required to reach a target \acrshort{qor}.
Speedup is the ratio of a method's speed to the baseline speed.
Fig. \ref{fig:speedup_combined}a shows that by using the statistical model, equivalent \acrshort{qor} is achieved in fewer runs on average.
This speedup is more pronounced for lower target values (higher \acrshort{qor}).
Table \ref{tab:prm_prediction} shows the speedup reached at a target transistor count of 365.
Furthermore, without \acrshort{iism}, circuits with the lowest transistor count could only be obtained with the prediction model approach.

\textbf{Two-Step Ahead Predictions (2SA):}
We further extend our method to 2\acrshort{sa} predictions, where the \acrshort{prm} and \acrshort{pom} select the \textit{pair} of recipes for the two next steps.
Table \ref{tab:prm_prediction} and Fig. \ref{fig:speedup_combined} show that 2\acrshort{sa} achieves drastically better speedups than 1\acrshort{sa}.
For a target transistor count of 360, 2\acrshort{sa} prediction reaches up to $14\times$ speedup.

\begin{figure}
    \centering
    \includegraphics[width=0.9\linewidth]{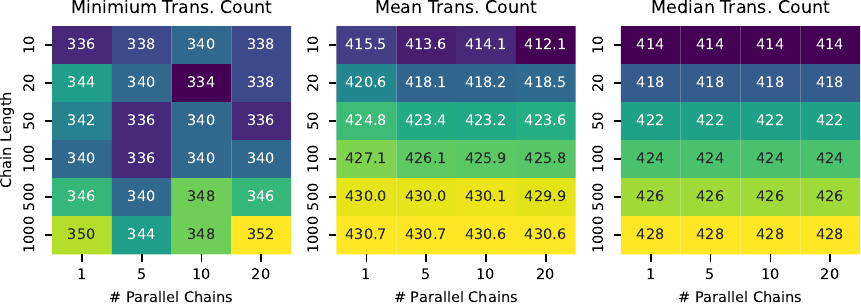}
    \caption{Effect of chain length and number of parallel chains on achieved minimum, mean and median transistor counts.}
    \label{fig:iism_heatmap}
    \vspace{-15pt}
\end{figure}

\textbf{\acrshort{iism} Effect:}
We compare 1\acrshort{sa} prediction with and without \acrshort{iism} in Table \ref{tab:prm_prediction}.
With chain length 50 and 1 parallel chain, we observe the speedup increases from 5.48 to 10.60.
We also perform a grid search over the number of parallel chains and chain length with uniform sampling (see Fig. \ref{fig:iism_heatmap}).
For this experiment, we keep the total number of steps per configuration constant by varying the number of runs.
We find that an increase in parallel chains is beneficial, which has been verified for different benchmark designs.
However, the optimal chain length varies widely across designs.
Also, while a configuration maximizing the \acrshort{qor} across many steps usually achieves high-quality, it does not always yield the absolute optimum (338 transistors count vs. absolute minimum of 334 in Fig. \ref{fig:iism_heatmap}).

\section{Results}
In the following, we deploy the EPFL Combinational Benchmark \cite{epfl_benchmark}.
Starting with the original Verilog implementation, we apply our method targeting \acrshort{mig} nodes minimization, and Table \ref{tab:benchmark} presents the resulting minimal circuit.
We demonstrate an improvement of up to $20.94\%$ in MIG minimization on the \texttt{max.v} circuit.
As the duration of each step grows dramatically with the design size (from 5\,ms to 21\,min for some recipes), we have not yet applied the full method on larger designs.
\acrfull{aig} node counts of the \acrshort{mig} optimized circuits are added for reference, but have not been optimized for.
The synthesized designs are available on \href{https://github.com/MaxenceBouvier/mig_minimized_designs_with_predictor_guided_random_dse}{GitHub}.

\begin{table}[htbp]
\centering
\caption{Best \acrshort{qor} achieved on the EPFL Benchmark\cite{epfl_benchmark}}
\label{tab:benchmark}
\begin{tabular}{l|rrr|r|rr}
\toprule
\multirow{2}{*}{\textbf{Bench.}} & \multicolumn{3}{c}{\textbf{MIG Nodes}} & \textbf{Depth} &\multicolumn{2}{c}{\textbf{AIG Nodes}} \\
 & \cite{date24} & \cite{dac24_mt}  &  Ours  & Ours & \cite{date19} & Ours  \\
\midrule
adder&384&384&384&129&&384\\
bar&2433&1906&\textbf{1800}&15&&2696\\
div&12462&\textbf{12368}&12434&2316&19250&23012\\
hyp&115541&\textbf{115539}&131487&8991&209460&235073\\
log2&22010&\textbf{22008}&23291&217&30522&30247\\
max&2190&1939&\textbf{1533}&260&&2676\\
multiplier&17112&\textbf{17112}&18515&141&25371&26381\\
sin&3870&3869&\textbf{3842}&118&4987&5102\\
sqrt&12357&\textbf{12247}&19430&5722&19706&19020\\
square&8138&\textbf{8089}&8132&128&17010&17838\\
arbiter&6711&792&\textbf{685}&36&879&783\\
cavlc&492&374&\textbf{329}&15&483&415\\
ctrl&74&60&\textbf{58}&9&&71\\
dec&304&304&304&3&&304\\
i2c&871&636&\textbf{618}&21&710&755\\
int2float&172&115&\textbf{107}&14&&145\\
mem\_ctrl&32097&6886&\textbf{6250}&36&7644&7102\\
priority&406&337&\textbf{315}&14&&375\\
router&147&97&\textbf{92}&14&96&95\\
voter&4555&3894&\textbf{3849}&41&9817&8854\\
\bottomrule
\end{tabular}
\end{table}
\vspace{-8pt}

\section{Conclusion}
The proposed method shows a 14$\times$ synthesis speedup and up to 20.94\% better MIG minimization through the integration of predictive models and iterative selection strategies.
Hyperparameter optimization remains crucial, as the best configuration varies among circuits.
Since a considerable number of steps are required to achieve the best \acrshort{qor}, minimization of large designs remains challenging.
We believe that further \acrshort{qor} improvement can be achieved with reinforcement learning methods.
As the \acrshort{prm}, \acrshort{pom}, and \acrshort{iism} are originally independent of the \acrshort{mt} software, our method could be deployed with other tools.


\begin{thebibliography}{00}
\bibitem{yosys} C. Wolf, “Yosys Open SYnthesis Suite.” \href{https://yosyshq.net/yosys}{Online}. 
\bibitem{abc_tool} R. Brayton and A. Mishchenko, “ABC: An Academic Industrial-Strength Verification Tool,” in Computer Aided Verification, 2010. \href{https://doi.org/10.1007/978-3-642-14295-6_5}{doi}.
\bibitem{mt_tool} H. Riener et al., “Scalable Generic Logic Synthesis: One Approach to Rule Them All,” in DAC, 2019. \href{https://10.1145/3316781.3317905}{doi}.
\bibitem{dac24_mt} S.-Y. Lee et al., “Late Breaking Results: Majority-Inverter Graph Minimization by Design Space Exploration,” in DAC, 2024. \href{https://doi.org/10.1145/3649329.3663507}{doi}.
\bibitem{predictive} F. Faez et al., “MTLSO: A Multi-Task Learning Approach for Logic Synthesis Optimization,” in Proceedings of the ASPDAC, 2024.
\bibitem{transformer} A. Vaswani et al., “Attention Is All You Need,” 2023 (v7). \href{https://arxiv.org/abs/1706.03762}{arXiv}.
\bibitem{epfl_benchmark} Luca Amarú, et al., "The EPFL combinational benchmark suite," in Proceedings of the IWLS, 2015.
\bibitem{date24} A. Tempia Calvino et al., "Scalable Logic Rewriting Using Don’t Cares," in DATE, 2024.
\bibitem{date19} E. Testa et al., "Scalable Boolean Methods in a Modern Synthesis Flow," in DATE, 2019.
\end{thebibliography}
\end{document}